\documentclass{article}

\usepackage{arxiv}

\usepackage[utf8]{inputenc} 
\usepackage[T1]{fontenc}    
\usepackage{hyperref}       
\usepackage{url}            
\usepackage{booktabs}       
\usepackage{amsfonts}       
\usepackage{nicefrac}       
\usepackage{microtype}      
\usepackage{lipsum}
\usepackage{graphicx}
\usepackage{amsmath}

\usepackage{caption} 

\renewcommand{\figureautorefname}{Figure~\negthinspace}

\title{Dynamic Deep Convolutional Candlestick Learner}

\author{
  Jun-Hao Chen\\
  Department of Computer Science and\\
  Information Engineering\\
  National Taiwan University\\
   \And
 Yun-Cheng Tsai\\
  Department of Data Science\\
  Soochow University\\
  \texttt{pecutsai@gm.scu.edu.tw} \\
}

\begin{document}
\maketitle

\begin{abstract}
Candlestick pattern is one of the most fundamental and valuable graphical tools in financial trading that supports traders observing the current market conditions to make the proper decision. This task has a long history and, most of the time, human experts. Recently, efforts have been made to automatically classify these patterns with the deep learning models. The GAF-CNN model is a well-suited way to imitate how human traders capture the candlestick pattern by integrating spatial features visually. However, with the great potential of the GAF encoding, this classification task can be extended to a more complicated object detection level. This work presents an innovative integration of modern object detection techniques and GAF time-series encoding on candlestick pattern tasks. We make crucial modifications to the representative yet straightforward YOLO version 1 model based on our time-series encoding method and the property of such data type. Powered by the deep neural networks and the unique architectural design, the proposed model performs pretty well in candlestick classification and location recognition. The results show tremendous potential in applying modern object detection techniques on time-series tasks in a real-time manner.
\end{abstract}


\section{Introduction}
\label{sec:introduction}
Object detection plays one of the central roles in computer vision. It has a wide variety of applications~\cite{liu2020deep, lin2017feature, redmon2016you}. Object detection tasks need to classify each object and precisely recognize their locations within the image. The successful implementation of such functions may benefit different fields in computer vision. There are several candidates, and artificial neural networks (ANN) is one of them~\cite{liu2020deep}. Artificial neural networks are universal function approximators. A single hidden layer neural network should be able to approximate any computable function~\cite{hornik1989multilayer}. However, the width of this hidden layer may exceed the limited computational resources and make it impractical. The development of convolutional neural networks (CNN)~\cite{krizhevsky2012imagenet, szegedy2015going, simonyan2014very, lecun2015deep, lecun1998gradient, goodfellow2016deep} revolutionized computer vision and the whole machine learning community. With CNN, it is possible to design deep neural networks capable of capturing the spatial relationships in the data and are with human-level or superhuman performances in classification accuracies.
Moreover, several variants of CNN, such as RCNN~\cite{girshick2014rich} and YOLO~\cite{redmon2016you}, begin to show up since $2014$ and bring us new possibilities in the application of CNN. The latest refined deep learning-based models still keep outperforming the state-of-the-art even now, not to mention the flourishing versatile applications in the real world. Impressed by the performance of object detection models, in this work, we aim to harness the strength of location recognition to time-series-related tasks in the financial field.
Time-series data is a sequence of data points indexed in the temporal order. It is a widely used data type in domains like the financial and medical industries. Meanwhile, time-series data are continuously accumulating in these sectors, creating an urgent need for well-designed machine learning models to analyze them. It is highly desirable in the financial industry if there is a model that can predict the future based on previous data. For example, in the stock market, past patterns are commonly used to predict potential future outcomes. Hence, the classification task plays a crucial role in time-series analytics. Our previous GAF-CNN model with a simple structure could quickly solve the multi-variate time-series classification problem in financial candlestick data~\cite{tsai2019encoding}. Nevertheless, the classification model is limited to predicting the classes based on fixed windows size inputs. 

The candlestick data is a specific genre of multi-variate time-series data that simultaneously contains open, high, low, and close time series. In financial trading, the candlestick is a frequently used tool supporting traders to observe the financial market pattern and make decisions. To automatically capture the candlestick pattern under different time scales, the model must classify the classes and recognize the locations according to traders' practical needs. Both demands, especially the latter, can potentially be solved by the strength of object detection models mentioned earlier.
To harness the power of the modern object detection deep learning models, we need to process the raw time-series data to fit CNN's input format first. A simple way to achieve this is to use time-series encoding methods converting one-dimensional time-series data into a two-dimensional matrix. Gramian Angular Field (GAF) is one of the encoding techniques which represents time-series data with polar coordinates and converts them into the symmetric matrix~\cite{wang2015imaging}. Due to its simple operations and rich information, we use GAF as our time-series preprocessing method.

The next thing to consider with the input data encoded as a matrix is the object detection model. The object detection model based on deep learning mainly contains two genres: (1) one-stage and (2) two-stage. The former has the advantage of speed, while the latter is better in performance. To be convenient and straightforward, a one-stage YOLO-$v1$ model is in this study~\cite{redmon2016you}. The YOLO model is one of the most remarkable and admired object detection models consistently evolving since $2014$. Despite the recent development in more powerful models, we found that the simple original version $1$ is sufficiently competent in our purpose. Moreover, based on the GAF matrix and time-series use cases, several adjustments will be applied to make the model more concise and practical in time-series applications.

We create a custom dataset for research regarding the data since there is no existing candlestick dataset for object detection tasks. we use foreign exchange EUR/USD $1$-minute data, from January $1$, $2000$, to January $1$, $2020$, as our raw data. As mentioned before, we label these data as eight candlestick pattern classes referring to the Major Candlestick Signals~\cite{MajorSignals}. We choose the top $10$ for each class as the targets. Next, we use these top 10 targets and apply the Dynamic Time Warping (DTW) method~\cite{berndt1994using} to collect the entire dataset with different window sizes. The processed final dataset contains eight candlestick pattern classes with window sizes from $5$ to $16$. More detail of our rule-based and data collection method is in Section~\ref{sec:background}.

Finally, we adopt both accuracy and confusion matrix to evaluate the performance of our candlestick detecting model. The innovative time-series object detection approach we proposed can reach around $88.35$\% classification accuracy and $75.55$\% accuracy on window sizes determination in the testing set. These results have sufficiently verified the object detection model. It is competent at time-series-related tasks and would pave the way to financial pattern detection in real-time scenarios.

\section{Background}
\label{sec:background}
\subsection{Candlestick Data}
\label{subsec:candlestick}
Entry and exit points detection play an essential role in financial trading. Crucial information for strategic decision-making often hides in the depths of unpredictable market prices. To capture this information more effectively, Munehisa Homma proposed an intuitive visualized tool, called \emph{Candlestick Chart}, to better observe the rice market's trading circumstances in the $18$th century~\cite{nison2001japanese}. A Candlestick chart is a visual representation that contains four-dimensional time-series data, the open, high, low, and close prices (OHLC) for a specific period. The period can be arbitrarily customized, usually depending on the transaction strategy. The OHLC data is defined as follows:

\begin{enumerate}
    \item Open: the first price of a specific period.
    \item High: the highest price of a specific period.
    \item Low: the lowest price of a specific period.
    \item Close: the last price of a specific period.
\end{enumerate}

\begin{figure}[ht]
\centering
\includegraphics[scale=0.5]{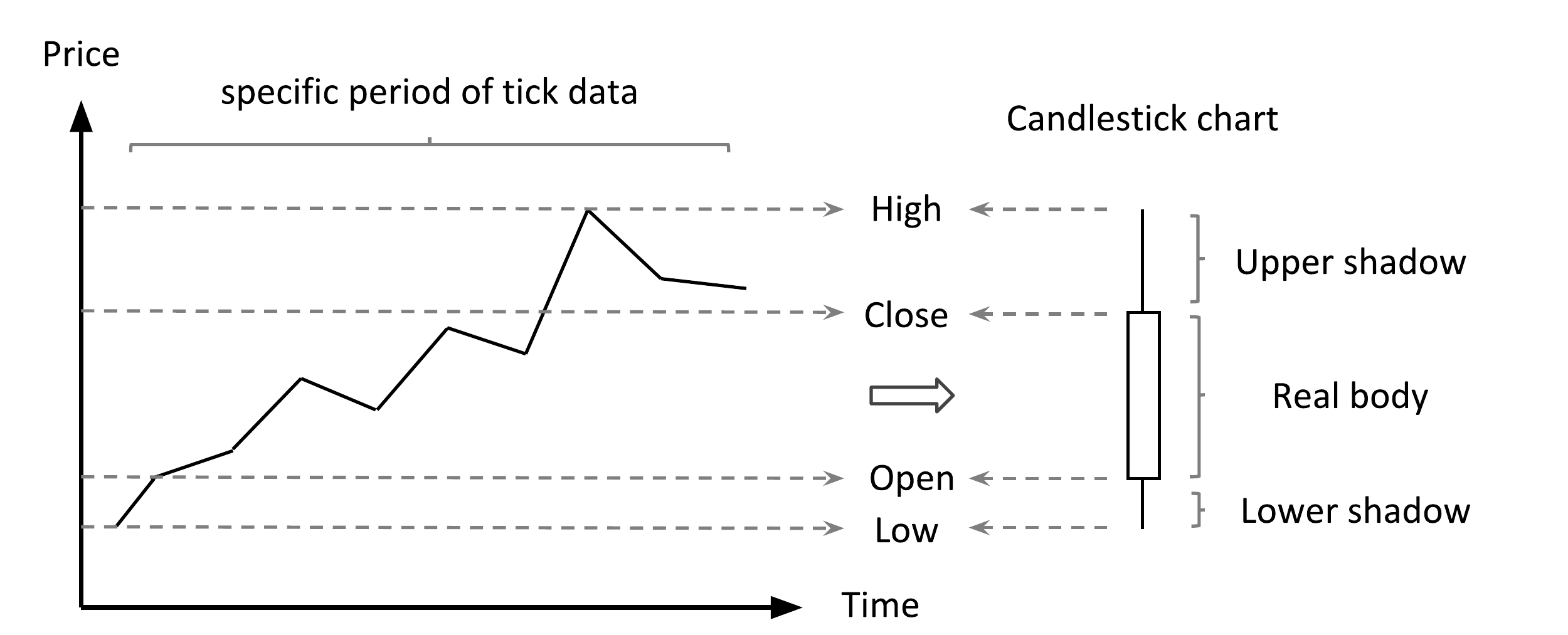}
\caption{Candlestick chart provides useful information for trading, such as open, high, low, and close prices.}
\label{candlestick_intro}
\end{figure}

Apart from the original OHLC data, \emph{Real Body} and \emph{Shadows}, operated by OHLC, are also practical features helping traders to make trading judgments well. The definitions are listed below. The real body will be colorized with black or red if the open price is higher than the close price. Otherwise, it will be colorized with white or green. The terminologies above are commonly used in professional traders’ discussions. \figureautorefname{ \ref{candlestick_intro}} illustrates the components of the candlestick chart in detail.

\begin{enumerate}
    \item Real Body: the length between the open and close price of a specific period.
    \item Upper Shadow: the length between the high price and $max(open, close)$ of a specific period.
    \item Lower Shadow: the length between the low price and $min(open, close)$ of a specific period.
\end{enumerate}

To make profits from financial trading, one of the critical skills is reading out the current market status from the candlestick data. It is challenging yet essential to extract meaningful information from candlestick data. There are certain kinds of \emph{candlestick pattern} which provide valuable information about the dynamics of the market. 
%
The candlestick pattern can help capture the critical circumstances among the supply and demand behind the financial market~\cite{lee2006pattern}. For a detailed description of the basic candlestick patterns, we refer to the book \emph{The Major Candlesticks Signals}~\cite{MajorSignals}. Each pattern introduced in the book is composed of two parts: 

\begin{enumerate}
    \item Trend pattern: A specific pattern within the prices of a period. The uptrend and downtrend are used in the reference to depict a pattern.
    \item Three bars pattern: The significant characteristics of the pattern appear in the last three bars/candles, which are crucial for human traders to make decisions.
\end{enumerate}

Besides, referring to~\cite{MajorSignals}, eight of these patterns are adopted in this study, including the Morning Star, Evening Star, Bullish Engulfing, Bearish Engulfing, Shooting Star, Inverted Hammer, Bullish Harami, and Bearish Harami patterns. Each of them will be well-illustrated in the Appendix.

\subsection{GAF Encoding}
\label{subsec:gaf}
Gramian Angular Field (GAF) is a time-series encoding method proposed by Wang and Oates~\cite{wang2015imaging}. The process encodes one-dimensional time-series data into a two-dimensional symmetric matrix with the matrix elements defined on the time-series represented in polar coordinates. The process contains three steps. First, normalize the raw time-series data $X$ to $[0, 1]$ with a simple Min-max normalization as equation \ref{equ:minmax_n}, where notation $\widetilde{x}_i$ denotes each normalized time-series in the entire normalized dataset $\widetilde{X}$.

\begin{align}
\widetilde{x}_{i}&=\frac{x_i-\min(X)}{\max(X)-\min(X)}
\label{equ:minmax_n}
\end{align}

Second, represent $\widetilde{x}_{i}$ into a polar coordinate system and calculate the cosine angle and radius with equation \ref{equ:gaf_arccos}. After this step, all of the normalized time-series data will be transformed into the angle representation with the arccosine function.

\begin{equation}
\begin{aligned}
\phi &= \arccos(\widetilde{x}_i), 0\leq \widetilde{x}_i \leq 1, \widetilde{x}_i \in \widetilde{X}\\
r &= \frac{t_i}{N}, t_i\in\mathbb{N}
\label{equ:gaf_arccos}
\end{aligned}
\end{equation}

Finally, use the cosine function to construct the symmetric matrix with the combination of the angles $ \phi_{i} + \phi_{j}$. The elements in the GAF matrix contain all permutations of any two tips of time-series data. In other words, the entire GAF matrix includes rich information of temporal correlations of each two data points. Overall, the processed final dataset contains eight candlestick pattern classes with window sizes from 5 to 16.

\begin{equation}
\begin{aligned}
\textup{GAF} &=\cos(\phi_i + \phi_j) \\ &=
\widetilde{X}^T \cdot \widetilde{X} - \sqrt{I-\widetilde{X}^2}^T\cdot \sqrt{I-\widetilde{X}^2}
\label{equ:gaf}
\end{aligned}
\end{equation}

After the encoding process, it can be found out that the diagonal elements of the GAF matrix represent the encoded information of original time-series data. The order of the timeline starts from the top left to the bottom right in the GAF matrix. Moreover, the non-diagonal elements represent the correlation between the two time-series data encoded information. In addition, the GAF matrix also contains an essential property. Mapping the normalized time-series data to the GAF matrix is bijective only if $\phi \in [0,\pi]$. This allows the GAF matrix to have one and only one result of time-series data with the unique inverse transformation through its diagonal elements.

\subsection{GAF-CNN in Candlestick Classification}
\label{subsec:gaf-cnn}
The GAF-CNN is literally a two-step process to train the CNN model with time-series data. However, before harnessing the strength of CNN models, time-series data need to be encoded into matrix format. Our previous study shows two valuable features to encode candlestick patterns:  (1) open, high, low, and close (OHLC) and (2) close, upper shadow, lower shadow, and real-body (CULR). Firstly, the OHLC features are just the raw candlestick data representation, as Section \ref{subsec:candlestick} mentions. And then, the CULR features are inspired by the observation habit of human traders. Human traders are used to capturing the subtle features through visual length rather than numerical value. Both parts perform well under a simple Lenet-5 structure. The following Figure \ref{fig:eve_ohlc_culr} is a GAF encoding example of both features of the evening star pattern.

\begin{figure}[ht]
\centering
\includegraphics[scale=0.5]{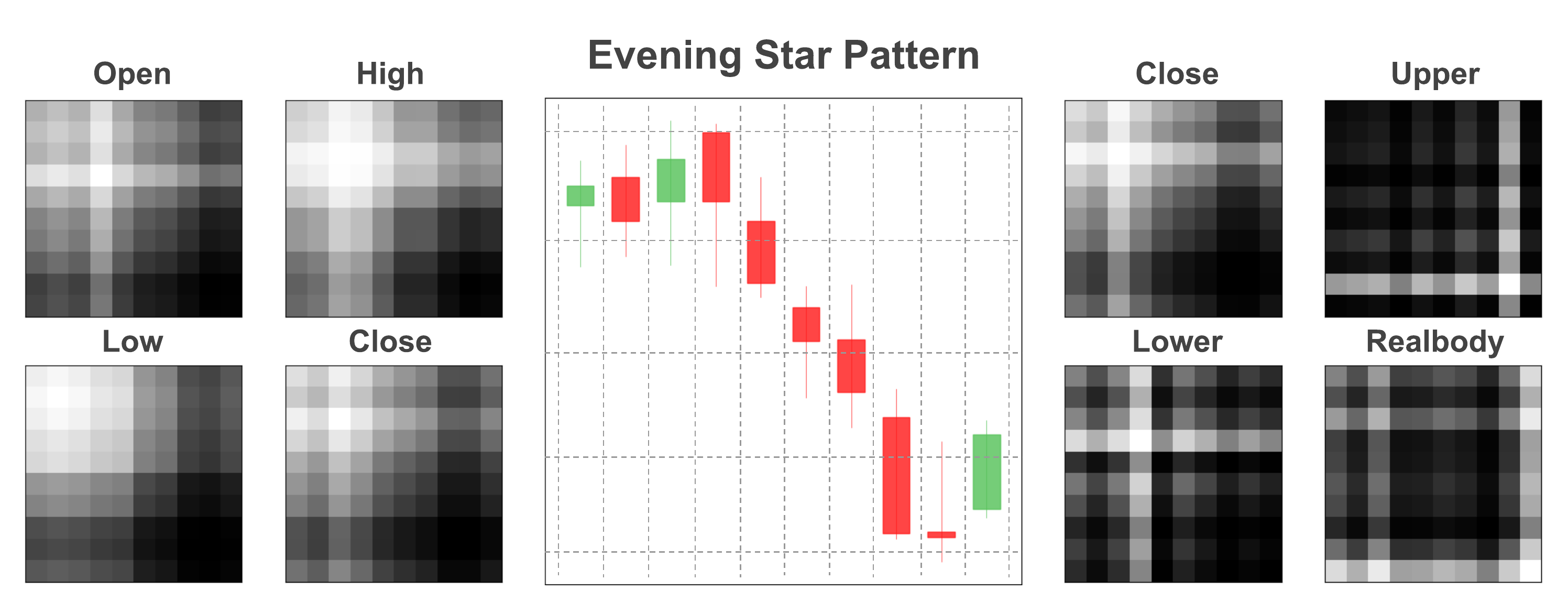}
\caption{A GAF encoding example of OHLC and CULR features of the evening star pattern. The left-hand side shows the OHLC features, and the right-hand side is the CULR features.}
\label{fig:eve_ohlc_culr}
\end{figure}

However, the length of window sizes is fixed in the simple classification task. It cannot fit more general financial applications—for example, different time scale detection. Besides, GAF encoding has excellent potential in integrating deep learning object detection techniques to develop a dynamic window length detection model. Therefore, the rest of this study will present a method to realize object detection with GAF time-series encoding.

\section{Data Preparation}
\label{sec:data_preparation}
Since there is no existing object detection dataset for the candlestick data, we need to create a custom dataset for the research. A three-step rule-based process is employed to prepare the dataset used in this study. First, we label the eight candlestick patterns according to the book \emph{The Major Candlesticks Signals}~\cite{MajorSignals} and target each class's top $10$ patterns. According to the abstract depiction of the eight candlestick patterns in the book, we handmade both rules of (1) trend pattern and (2) three bars pattern rules with our handmade conditions. The trend pattern is defined as the top $20$ percentiles of positive and negative slopes. The value below the $20$ percentile will regard as no trend existing.
Furthermore, we choose the patterns with the top $10$ steep slopes as the targets, including ten positive and ten negative slopes. Additionally, the three bars pattern definitions are more complex than the trend pattern; therefore, we present an Evening Star case as an example below. More details of the rules will show in our released code.

\begin{figure}[ht]
\centering
\includegraphics[scale=0.45]{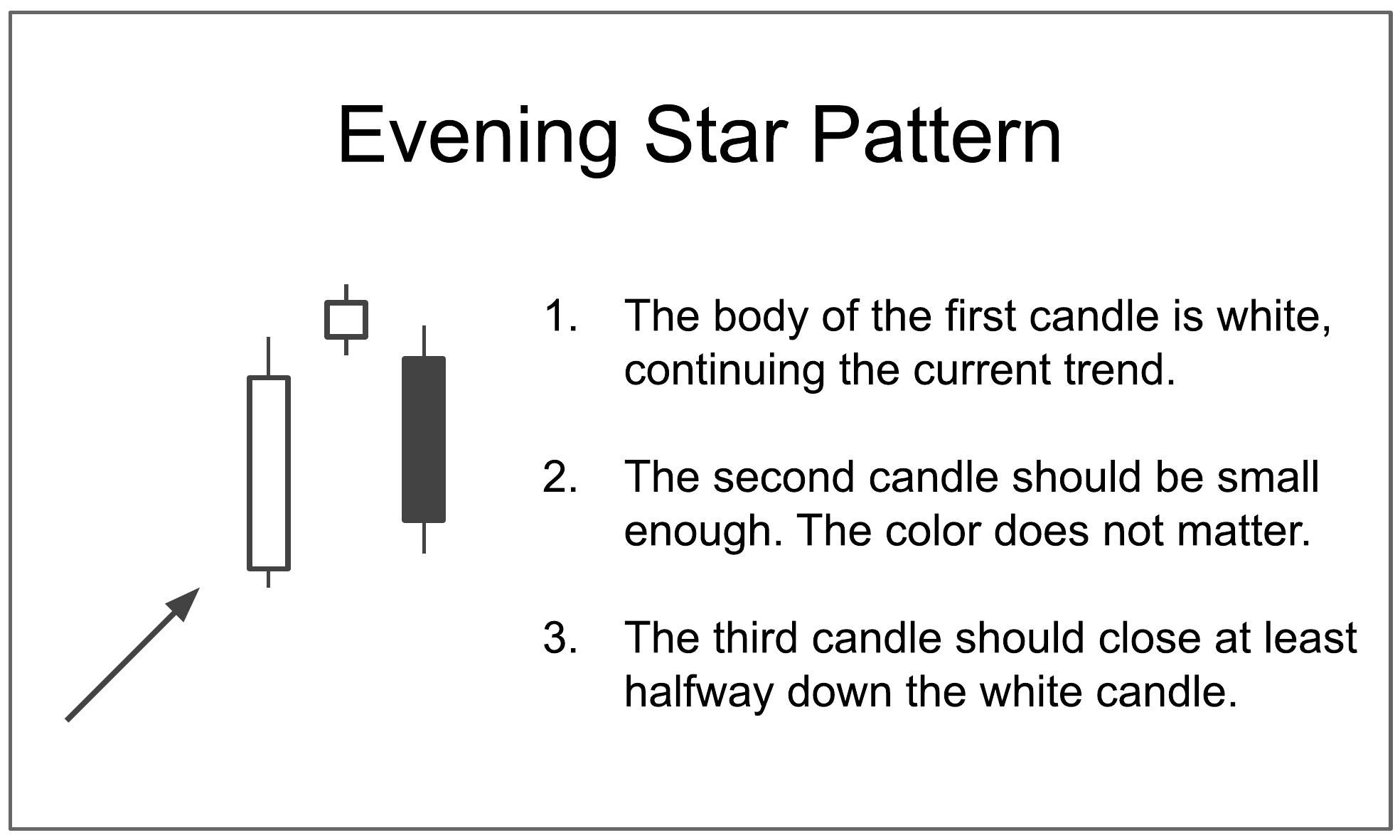}
\caption{The three bars pattern depiction of the Evening Star pattern. The textual description illustrates the three bars pattern. The trend pattern is only present by the arrow.}
\label{evening_star}
\end{figure}

In \figureautorefname{ \ref{evening_star}}, several essential required rules are depicted to form an Evening Star pattern, and most of the conditions can rule literally. For instance, the first condition requires the bar/candle's body to be white so that the rule will be the (close - open) value of the first bar should be greater than zero. However, not all depiction can rule quickly; some of the descriptions, like "enough," is very abstract, which is also very subjective to different traders. This study uses the value greater than $75$ and lower than $25$ percentile to represent the "enough" case. Still, the actual application is supposed to label with traders themselves.

Second, we then apply the Dynamic Time Warping~\cite{muller2007dynamic} method to compare the raw foreign exchange data. Each class's top $10$ patterns only collect the data if it is lower than the DTW distance threshold. Here, the DTW distance means the similarity of the compared data to the top $10$ patterns. The $20$ percentile of the distance can collect enough data to verify our idea. After this step, we will get plenty of similar patterns to the top $10$. However, the data still lacks the clear and definite window sizes to be detected by the model.

\begin{figure}[ht]
\centering
\includegraphics[scale=0.45]{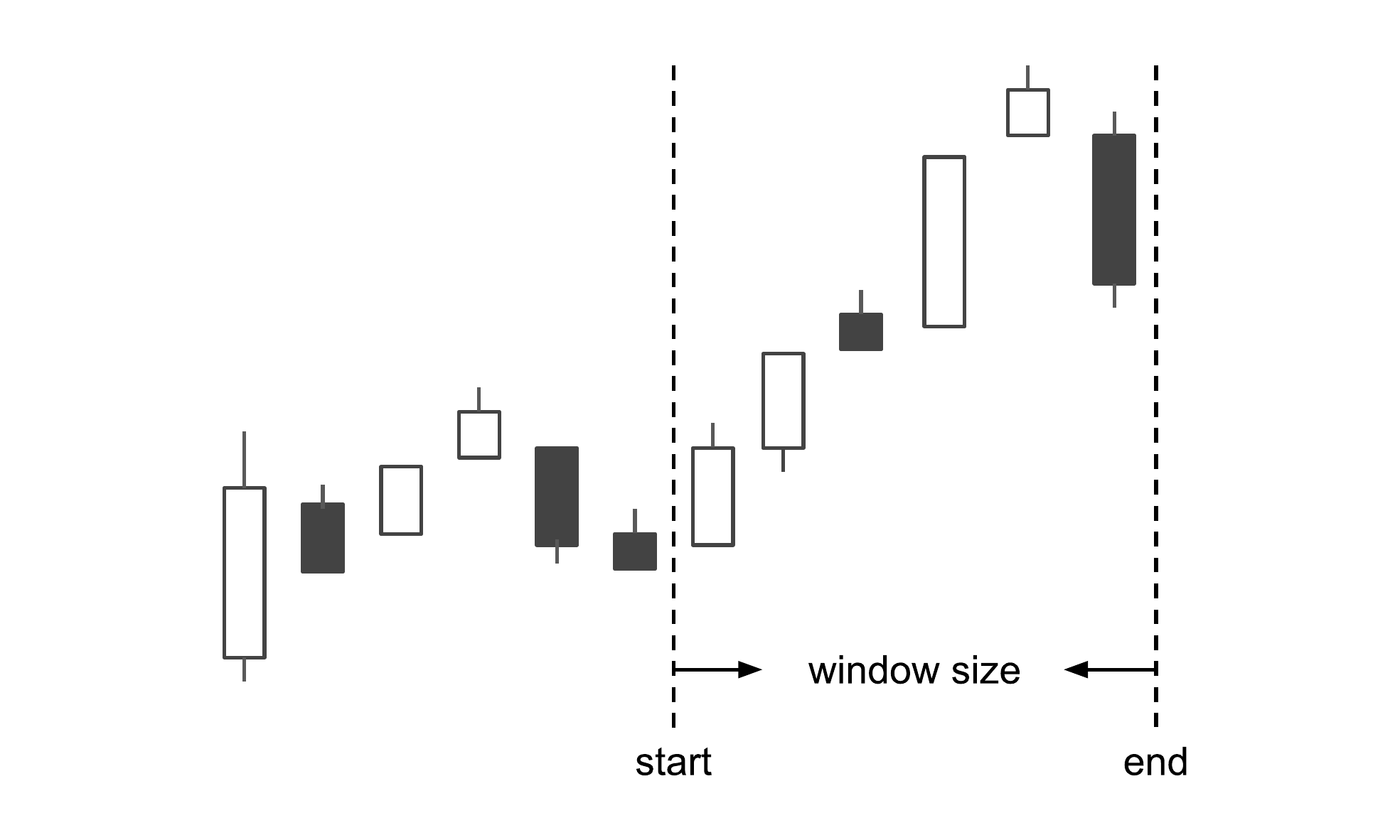}
\caption{The illustration of the window size. According to the real-time usage characteristic of time-series data, we only need to define the start of the pattern.}
\label{cuttail}
\end{figure}

The rule-based process's last step is to define these window sizes precisely. Reasonably, we assume that traders can recognize their patterns' start and the end. In the language of object detection, this means that the bounding box's position should be well-defined and not an ambiguous label. Hence, we adopt the price reversion of trend representing the patterns' start so that the window size will be the distance between the start and the end of each pattern. The price reversion is defined as the candlestick's color change in contrast to the trend after the $5th$ bar. For example, \figureautorefname{ \ref{cuttail}} is the case of the Evening Star pattern containing a positive trend. Therefore, the price reversion here means the color changes from white to black. Based on the difficulty of collecting real data, our rule-based process simulates traders' labeled patterns. All rule-based data are required to satisfy the strictness and consistency among the same class and window size.

\section{Methods}
\label{sec:methods}
The typical design of the YOLO-$v1$ model is detecting objects in the visual images, such as Microsoft COCO or CIFAR-$10$ dataset. However, it is a significantly different data form in the case of the time-series GAF encoding matrix. Moreover, according to the real-time usage property of time-series data, the bounding box can also be simplified. Hence, we make the model's predictions more concise and compelling in the case of the GAF matrix. We make three crucial modifications from the original model based on the properties of the GAF matrix and the real-time usage situation as below.

\begin{figure}[htbp]
\centering
\includegraphics[scale=0.6]{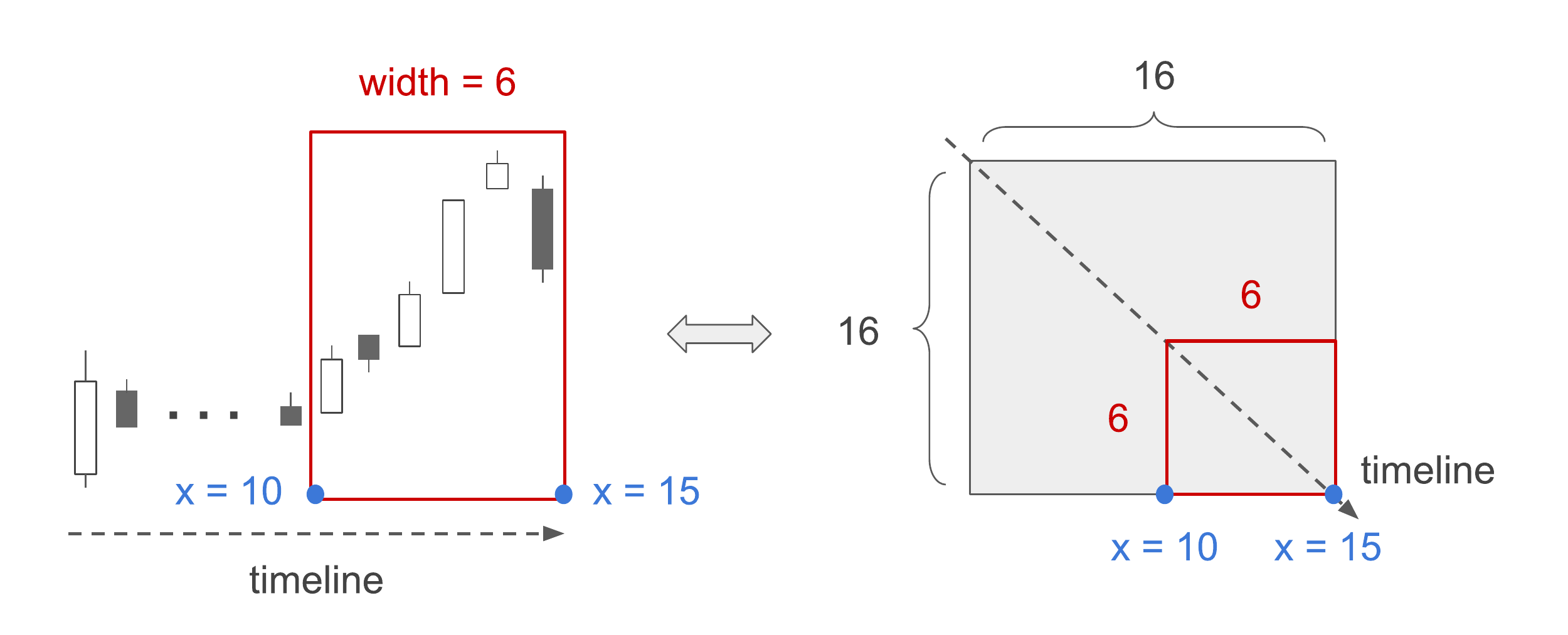}
\caption{As mentioned in Section~\ref{subsec:gaf}, the bottom right of the GAF matrix represents the latest time-series data. Instead of the original grid system, the start point of GAF matrix only needs to define the location $(x, y)$ at the bottom right corner. And the model will define the window size $w$.}
\label{fig:grid_system}
\end{figure}

\begin{enumerate}

    \item Symmetric matrix: 
    
    The original YOLO-$v1$ model adopts the bounding box with the shape as $(x, y, h, w)$. The notation $(x, y)$ represents the position of the $x$ and $y$-axis, and $(w, h)$ means the height and width predictions of the bounding box. Based on the property of the symmetric matrix, we only need the information of one side of the off-diagonal elements. In other words, we can remove the redundant side of the bounding box prediction; for example, we only need the information of $x$ and $h$. Therefore, the predictions $(x, y, h, w)$ of the bounding box in the original YOLO-$v1$ model can simplify to $(x, h)$. 
    
    \item Application scenario of time-series model: 
    
    The position $x$ in the YOLO-$v1$ model can be anywhere corresponding to the image, but it is not the same in the GAF matrix case. The moving-window framework needs to be adopted in a real-time usage situation. It needs to keep inferring the latest results and constantly to update time-series data, especially in financial trading. With the moving-window framework, we do not need to localize the object's position. The proper location should always fix at the latest time-series data point, which is the bottom right corner of the GAF matrix. For instance, the position $x$ should always be $16$ under a $16 \times 16$ size GAF matrix. \figureautorefname{ \ref{fig:grid_system}} illustrates the specific position.  In \figureautorefname{ \ref{fig:grid_system}}, we use a bounding box with a window size equal to 16 as an instance. The candlestick data form is on the left-hand side with the timeline from left to right. The corresponding GAF matrix with the timeline from top-left to bottom-right is on the right-hand side. As the timeline shows, $x$ equals $15$ is the latest position of this time-series data. With the rightest side as the starting point, the object can be framed by the width based on the above-mentioned symmetric property.
    
    \item Remaining modification: 
    
    The original grid system in the YOLO-$v1$ model divides the input image into an $S \times S$ grid, where $S = 7$ in YOLO-$v1$, but the grid system needs to be modified to correspond to the GAF matrix. In this research, we are building a model to output candlestick pattern labels and the size of the time window. What we consider is only the length counting from the bottom-right corner. We do not consider other starting points. Hence, we can simplify the grid system from $S \times S$ to $1 \times 1$. In other words, the grid system is not needed in our time-series moving-window framework. \figureautorefname{ \ref{fig:grid_system}} also helps to figure out this point.
\end{enumerate}

\begin{figure}[htbp]
\centering
\includegraphics[width=1.0\textwidth]{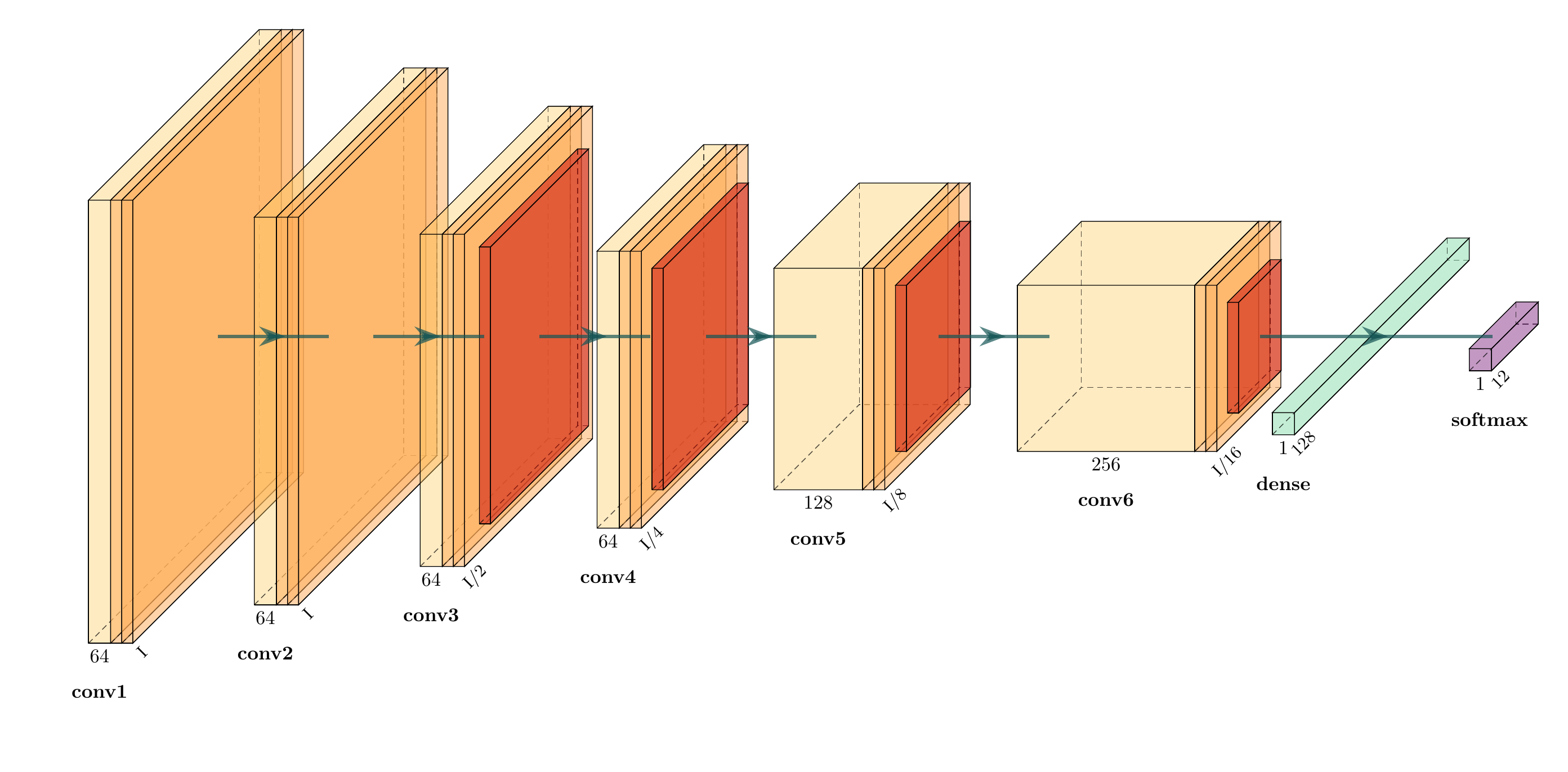}
\caption{The structure of our YOLO-$v1$ model. The modified YOLO-$v1$ model includes six $conv$-layers, followed by a dense-layer. Each $conv$-layer consists of a convolutional layer, a batch normalization layer, and a leaky-relu activation.}
\label{fig:yolomodel}
\end{figure}

\figureautorefname{ \ref{fig:yolomodel}} shows the entire architecture of our modified YOLO-$v1$ model. Each $conv$-layer contains a convolutional layer, a batch normalization layer, and a leaky-relu activation. Since input with $16 \times 16$ sizes, we design the max-pooling layer four times to make the feature map one-dimensional in the $conv6$-layer. We also reduce each layer's parameters to avoid using too many parameters in a more straightforward time-series task. With all the modifications, the last layer's predictions contain $12$ values: two pairs of $(w, c)$ and eight classes.

\section{Results}
\label{sec:results}
We train the modified time-series object detection model with the dataset labeled with eight classes and window size range from $5$ to $16$. The entire dataset is pre-split into training, validation, and testing sets with the $80/20/20$\% ratio. Besides, the process of splitting data is arranged to avoid using future information in time-series data. Detailed information about the experiment parameters setting in this work is shown in Table~\ref{tab:parameters}.

\begin{table}[htbp]
\centering
\begin{tabular}{|c|c|}
\hline
\textbf{Parameters} & \textbf{Values}   \\ \hline
optimizer           & adadelta          \\ \hline
learning rate       & 0.001             \\ \hline
rho                 & 0.95              \\ \hline
decay               & 0.0005            \\ \hline
epochs              & 4000                \\ \hline
\end{tabular}
\caption{The parameters used in our training process.}
\label{tab:parameters}
\end{table}

\begin{figure}[htbp]
\centering
\includegraphics[scale=0.7]{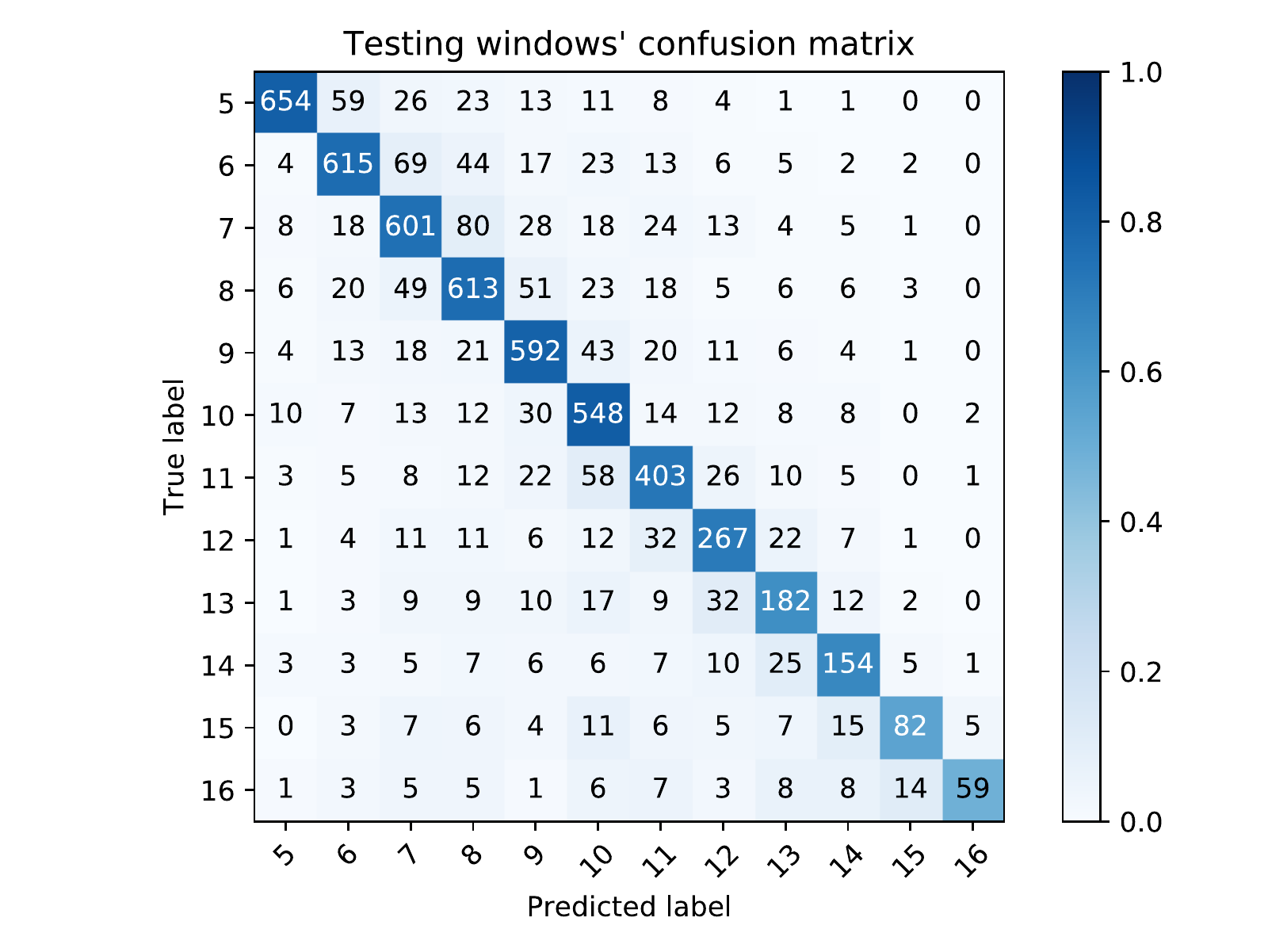}
\caption{The confusion matrix of testing on window sizes. The row and column represent the true label and predicted label, respectively. The background color represents the accuracy of each window label. The results show stable performance of the model on determining the window size.}
\label{test_win_cm}
\end{figure}

The testing results in Table~\ref{tab:classes_testing_result} show that the model can achieve at least $80.1$\% accuracy in each class. The overall classification accuracy reaches $88.35$\% on average. The average performance is very close to $90.7$\%, which is the original GAF-CNN model's performance solving only the simple classification problem~\cite{chen2020encoding}. Moreover, \figureautorefname{ \ref{test_win_cm}} shows the confusion matrix of testing on window size. The prediction results of the model are mostly on the diagonal, implying that the model performance is pretty well. Concretely, the window size prediction can achieve almost $75.55$\% correct on average in the testing set.

\begin{table}[htbp]
\centering
\begin{tabular}{ccc} \toprule
Label & Number of Samples & Accuracy (\%) \\
\midrule
1     & 789               & 98.9          \\
2     & 803               & 98.1          \\
3     & 855               & 90.3          \\
4     & 918               & 85.3          \\
5     & 641               & 82.4          \\
6     & 582               & 80.1          \\
7     & 892               & 82.8          \\
8     & 838               & 88.9          \\
\midrule
Avg   &                   & 88.35         \\
\bottomrule
\end{tabular}
\caption{The results of testing accuracies. The accuracy of each label means the one versus all accuracy. The accuracy on average is calculated by the arithmetic mean.}
\label{tab:classes_testing_result}
\end{table}

\begin{figure}[htbp]
\centering
\includegraphics[width=1.0\textwidth]{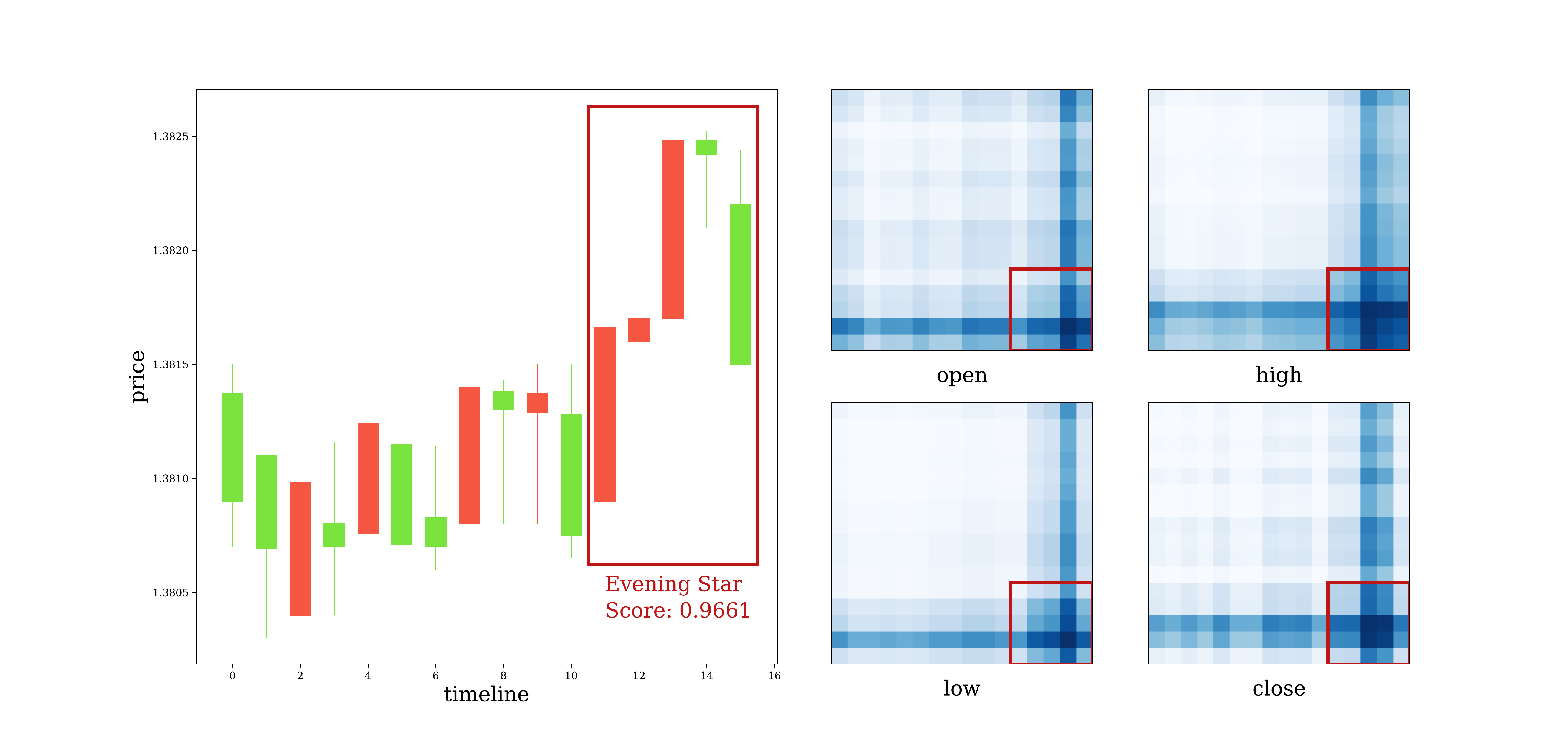}
\caption{An example of detection results for both candlestick chart and GAF matrices.}
\label{detection_1}
\end{figure}

\begin{figure}[htbp]
\centering
\includegraphics[width=1.0\textwidth]{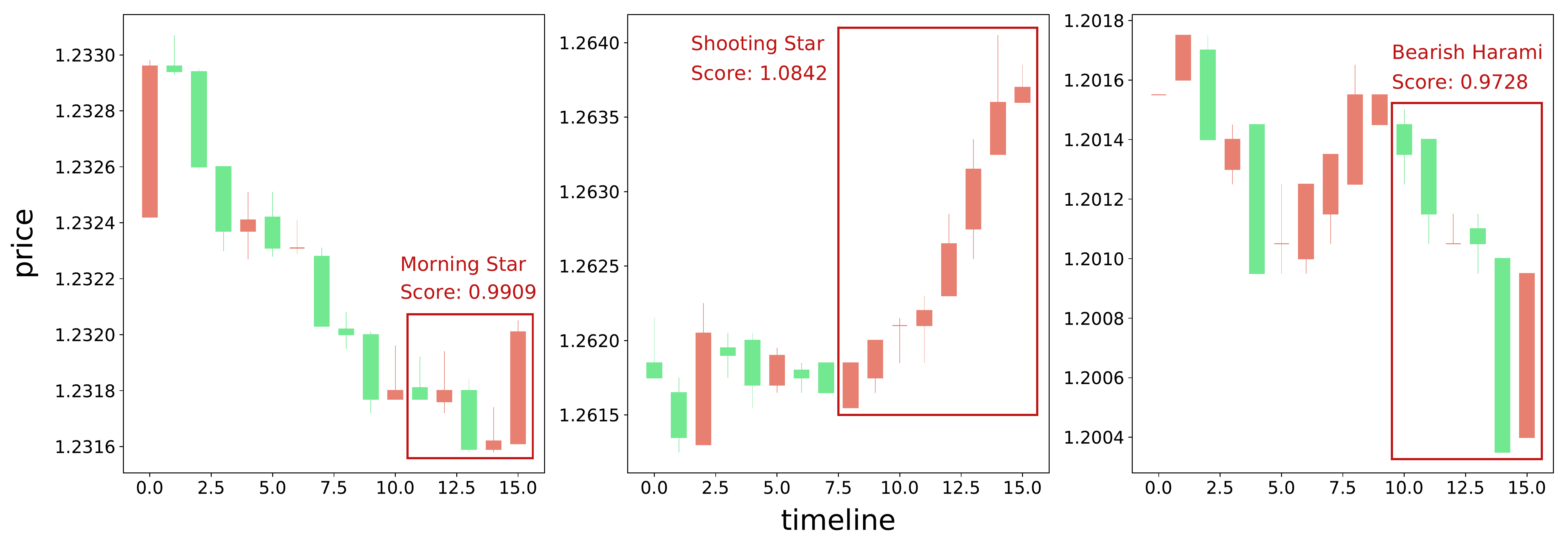}
\caption{Further examples of our detection results. The patterns' names and scores are labelled in each subfigure.}
\label{detection_2}
\end{figure}

Under the limited dataset, both the classification and location recognition results are optimistic. Based on the invertible transformation property of the GAF matrix, we can convert the detection results from GAF matrices back to the more intuitive candlestick chart. \figureautorefname{ \ref{detection_1}} presents a concrete detection example of the Evening Star pattern. The right-hand side is the bounding box of the stacked OHLC inputs, and the left-hand side is the candlesticks case. The predicted window size is labeled in the red box. \figureautorefname{ \ref{detection_2}} shows more examples of our detection results.

Our results show that the object detection model with simple architecture could solve both the classification and location recognition well in time-series data. It also helps detect patterns with dynamic length and different periods according to traders' practical needs. Furthermore, with the moving-window framework, real-time time-series object detection can easily be realized, benefiting professional traders in financial industries.  

\section{Discussions}
\label{sec:discussions}
\subsection{Advanced applications with modern object detection models}
As demonstrated above, the GAF encoding method integrates the time-series data with object detection models. It has superior potential for solving dynamic length detection of time-series data. In this work, we present an instance of the YOLO-$v1$ model. However, there are still numerous variants of modern object detection models that can be further applied. For example, the two-stage models like RCNN~\cite{girshick2014rich} may perform better inaccuracy, not to mention the state-of-the-art approaches. Besides, it is interesting to implement other computer vision techniques like semantic segmentation~\cite{chen2017deeplab} and instance segmentation~\cite{he2017mask} to the time-series domain. We leave these potential directions for future research.

\subsection{Robustness of dynamic models}
For a machine learning model to be confidently used in critical application scenarios, such models must be robust to adversarial noises. Deep learning models are known to be susceptible to such attacks~\cite{kurakin2016adversarial}.
In the field of objection detection, there are known attacks, for example~\cite{eykholt2018robust}. In the previous work~\cite{chen2020adversarial}, the authors studied the potential attacks on the deep candlestick learner, which operates on static inputs. We build the model on top of objection detection models in this work, which can process real-time data. It is interesting to study the effects of such attacks on our new model and the potential defense mechanisms.

\section{Conclusions}
\label{sec:conclusions}
This work presents an innovative computer vision application combining object detection and time-series data processing in the financial trading field. We consider the actual time-series application scenario and adjust to simplify the original YOLO-$v1$ model. With these crucial modifications, the modified time-series object detection model becomes more efficient and can apply in the real-time detecting framework. Even with limited labeled data and computing power, we can still achieve $88.35$\% classification accuracy and almost $75.55$\% window size prediction accuracy on average in the testing set. These results are very positive in both classification and location recognition. 
From the perspective of trading, patterns are generally too subjective to be captured by a general algorithm. However, there is tremendous potential in applying the object detection model to solve these emotional time-series problems. The help of expert labeling and the universal approximation capability of deep learning. We expect the proposed framework would benefit the financial sectors and other fields dealing with sophisticated time-series data.

\bibliographystyle{unsrt}  

\bibliography{references,bib/classical_nn,bib/classical_cnn,bib/adversarial}






\end{document}